\documentclass[preprint,12pt]{elsarticle}

\usepackage{amsmath}
\usepackage{amssymb}
\usepackage{graphicx}
\usepackage{booktabs}
\usepackage{longtable}
\usepackage{tabularx}
\newcommand{\hd}[2]{\begin{tabular}[b]{@{}l@{}}#1\\#2\end{tabular}}
\newcolumntype{L}[1]{>{\raggedright\arraybackslash\hsize=#1\hsize}X}
\usepackage[skip=6pt,labelfont=bf,labelsep=period]{caption}
\usepackage{array}
\usepackage{calc}
\usepackage{url}
\usepackage[T1]{fontenc}
\usepackage[utf8]{inputenc}

\journal{arXiv}

\providecommand{\tightlist}{\setlength{\itemsep}{0pt}\setlength{\parskip}{0pt}}

\begin{document}

\begin{frontmatter}

\title{From 80$\times$ to 385$\times$: A Best-Matching-Unit Search at the L2 Roof,
Measured Against a Symmetrically Tuned Baseline}

\author[jcu]{Andrew J. Amos}
\ead{andrew.amos@my.jcu.edu.au}
\ead[url]{https://orcid.org/0000-0002-9145-0212}
\address[jcu]{College of Medicine and Dentistry, James Cook University,
Townsville, Queensland, Australia}

\begin{abstract}
Comparisons between GPU implementations are usually asymmetric: one side is tuned by its
author, the other is run as found. I report a programme that tuned both a novel SOM
algorithm (SparseBin) and the baseline algorithm it was being compared to (cuSPARSE). The
best-matching-unit search that dominates self-organizing map training was tuned through
four levers -- tile size, tile-membership clustering, neuron-axis chunking and vectorised
loads -- reaching 5.6-10.1$\times$ per epoch over the previously published configuration at
map sizes from $32^2$ to $512^2$, and lifting the margin over the CUDA implementation
behind our earlier MEDLINE atlases from ${\sim}80\times$ to ${\sim}385\times$. The cuSPARSE
implementation SparseBin is compared to received every lever with an analogue on its side,
and became 2-3$\times$ faster in the process. The tuned kernel pressed the L2 bandwidth roof
at 77\% of peak with every other unit at 40-65\%, bounding any further lever at
${\sim}1.3\times$ -- a terminal result rather than a waypoint, and every untested lever was
either capped by that bound by construction or measured null.
\end{abstract}

\begin{keyword}
self-organizing map \sep GPU \sep CUDA \sep sparse matrix \sep kernel tuning \sep roofline
\end{keyword}

\end{frontmatter}

\section{Introduction}\label{introduction}

Every comparison between two machine learning algorithms has a
denominator, and it is common to see comparisons that only tune the
author's kernel. A true understanding of the relative merits of two
algorithms requires that both be tuned for peak performance. This paper
describes how symmetric tuning illuminated the comparison between
alternative algorithms for the best-matching-unit (BMU) search that
dominates self-organizing map training \cite{kohonen2013}.

At the scale of a MEDLINE corpus - 26.9 million abstracts as sparse
binary term vectors - the search is bound by the bandwidth needed to
read the codebook every epoch. A companion paper \cite{amos2026} shows that
storing the codebook feature-major, with each feature's weights
contiguous, recasts the search as a tiled sparse-dense product in which
every loaded weight column is reused across a tile of articles, and that
this accelerates the search 4.5-8.5$\times$ at no cost in quality, since an
exact-argmin BMU is invariant to how the codebook is stored \cite{amos2026}.
That paper compared its novel SOM algorithm against a baseline SOM
algorithm constructed with NVIDIA cuSPARSE library tools. Hereafter I
will refer to the novel SOM algorithm as SparseBin and the baseline
algorithm as cuSPARSE.

This article reports the symmetric tuning of both algorithms. Every
lever proven on one was offered to the other before any result was
reported, structural asymmetries were named rather than left as an
unexplained effort gap, predictions were registered before measurement,
and nulls were recorded at the same prominence as wins. Making cuSPARSE
faster counted as success - and it succeeded. The tuned cuSPARSE
reported here is 2-3$\times$ faster than the configuration in the original
paper \cite{amos2026}.

Four levers - tile size, tile-membership clustering, neuron-axis
chunking and vectorised loads - accelerated the BMU search 5.6-10.1$\times$ per
epoch over the previously published configuration, across map sizes from
32\textsuperscript{2} to 512\textsuperscript{2}. The margin over the CUDA implementation behind our earlier
MEDLINE atlases \cite{amos2024a,amos2024b} rose from \ensuremath{\sim}80$\times$ to
\ensuremath{\sim}385$\times$.

The tuned SparseBin kernel was the first configuration in the sequence
to press a hardware bandwidth ceiling rather than sit beneath all of
them: it reached 77\% of the L2 bandwidth roof while every other unit
ran at 40-65\%. Section 6 shows that every lever left on the shelf was
either capped by that bound by construction or measured null.

\subsection{Scope}\label{scope}

\begin{itemize}
\tightlist
\item
  All measures were completed on one RTX 4090. Every optimum here - tile
  size, chunk count, batch, the L2 roof finding itself - is tuned
  against that card's 72 MB L2 and register file. The first paper
  reported a 1024\textsuperscript{2} run on a 141 GB H200, but no tuning
  was done so it is not mentioned in this article.
\end{itemize}

\subsection{Contributions}\label{contributions}

\begin{enumerate}
\def\labelenumi{\arabic{enumi}.}
\tightlist
\item
  A measured 5.6-10.1$\times$ per-epoch speed-up of the BMU search, and an
  \ensuremath{\sim}80$\times$ $\rightarrow$ \ensuremath{\sim}385$\times$ margin over MedSOM (Section~5).
\item
  A terminal-limiter result: the tuned kernel reaches 77\% of the L2
  bandwidth roof, bounding further gains at \ensuremath{\sim}1.3$\times$, with
  every unexplored lever either capped by that bound by construction or
  measured null (Section~6).
\item
  A public artefact that re-derives the headline table from a fresh
  clone (Section~9).
\end{enumerate}

\section{Method: the rules the programme ran
under}\label{method-the-rules-the-programme-ran-under}

Four rules governed the programme. They were fixed before it began and
are stated here as a protocol another comparison could adopt, with what
each one cost.

\textbf{The objective was the best algorithm at each map size, not the
best configuration of one implementation.} Making cuSPARSE faster
counted as success.

\textbf{Parity was a rule, not an intention.} Every lever proven on one
side was offered to the other before either result was reported.

\textbf{Nulls were reported with the same prominence as wins.} Section 7
is the register. A lever that returned under 5\% at every map size was
recorded and abandoned rather than pursued.

\textbf{Predictions were registered before measurement.} Section 7 gives
every one with its outcome, including the failures.

\textbf{Quality was excluded from the search and restored at the end.}
Optimising per-epoch time alone made exhaustive search affordable: a
converged run costs minutes to hours, a one-epoch timing seconds.
Quality was recorded throughout but never used to select, then validated
on the winners in a separate phase.

\section{Background: the design being
tuned}\label{background-the-design-being-tuned}

This section states only what is needed to follow the tuning programme.
The layout argument, its proof of best-matching-unit invariance, and the
MEDLINE atlas it was built for are in the first paper \cite{amos2026}.

A self-organizing map is trained by repeatedly finding, for each
article, the neuron whose weight vector is closest to it - the
best-matching unit - and then updating that neuron and its neighbours.
On a corpus of sparse binary term vectors the search dominates: it was
98.7-99.5\% of epoch time before this programme began. It is bound not
by arithmetic but by the bandwidth needed to read the codebook every
epoch.

The paper's contribution is that this bound is largely an artefact of
how the codebook is stored for a sparse SOM. The SparseBin algorithm
stored it feature-major, with each feature's weights contiguous at
\texttt{W{[}v.M+i{]}} for feature \texttt{v} and neuron \texttt{i},
which recasts the search as a tiled sparse-dense product in which every
loaded weight column is reused across a tile of articles. Because an
exact-argmin best-matching unit does not depend on the order weights are
held in, the gain is free: held-out quantisation error agrees with
cuSPARSE to within 0.5\% at every map size. The kernel also fuses the
argmin into the product, so the score block is never materialised.

cuSPARSE, the comparison algorithm, is built on the proprietary NVIDIA
library of that name \cite{cusparse}. It computes the sparse-dense product with
a library call, materialises the resulting score block, and then reduces
it with a separate argmin pass. Three facts about it matter for the
parity rule in section 2. It can select among the library's algorithms;
it can reduce the precision of its score block; and its batch size sets
that block's footprint and therefore its cache behaviour. None of the
three has an analogue in a fused kernel, so they are structural
asymmetries rather than levers withheld.

\section{The levers}\label{the-levers}

Four levers produced the speed-up: tile size, tile-membership
clustering, neuron-axis chunking, and vectorised loads. They are
reported in the order they were run, because the sequence is what
identified the binding resource at each stage. A fifth lever, execution
ordering, was run and then made redundant by the clustering that
followed it; it is kept in the account because it is the experiment that
identified the mechanism the clustering went on to exploit.

\subsection{Tile size}\label{tile-size}

The kernel published in the first paper \cite{amos2026} processes the corpus in
tiles of 16 articles: for each tile it forms the union of the distinct
features those 16 articles touch, then sweeps that set of feature
columns once. The tile buys reuse and costs registers - five
sixteen-element per-article arrays held live per thread.

The tile size was swept across 2, 4, 8, 16 and 32 articles at every map
size. \textbf{Tile 16, the published value, was fastest at none of
them.} A tile of 2 was 2.7x faster at 128\textsuperscript{2}, and a tile of 8 was 1.5x
faster at 256\textsuperscript{2}.

The mechanism was not the one hypothesised. The expectation was that
smaller tiles would cut registers and raise occupancy while raising DRAM
traffic, since each loaded feature column is amortised over fewer
articles, producing an interior optimum between the two. DRAM traffic
turned out to be \textbf{flat in tile size}, within 4\% across the whole
sweep. The extra codebook re-reads of small tiles are absorbed by L2,
which rises from 6.5 to 9.7 TB, and L2 had headroom. The trade is
occupancy against \textbf{L2} traffic, not against DRAM, and occupancy
wins by a wide margin.

Two further properties emerged and both matter later. The optimum is
map-size dependent, so it is a variable rather than a constant. Changing
it also perturbs about one assignment in $10^5$ at 256\textsuperscript{2}, through
floating-point tie-breaking rather than any defect in the search.

\subsection{Locality: ordering, then membership, then
chunking}\label{locality-ordering-then-membership-then-chunking}

Three levers followed, each aimed at the L2 traffic the tile sweep had
identified as binding.

\textbf{Execution ordering.} The kernel launches one block per tile of
articles in a one-dimensional grid, so the blocks running at any given
moment are consecutive tiles in corpus order - which is accession order,
and therefore unrelated to what the documents contain. Each block had to
read the codebook columns for the features its own articles contained.
When two blocks running at the same time need many of the same features,
the first block's reads pull those columns into L2 and the second finds
them already there; when their features are disjoint, each block fetches
its own columns from DRAM.

Permuting which tiles run together, so that simultaneous blocks share as
many features as possible, cut DRAM traffic by 43\%, 40\% and 30\% at
64\textsuperscript{2}, 128\textsuperscript{2} and 256\textsuperscript{2} at the smallest tile, and raised the L2 hit rate from
49.5\% to 64.6\% at the largest. Which articles sit in which tile is
unchanged, so the order terms are summed in is unchanged, and
assignments are bit-identical by construction: twelve of twelve byte
comparisons matched.

The lever is conditional rather than universal. At a tile of 16 each
block already spans sixteen articles, so its own set of features is
large and overlaps its neighbours' heavily without any help; there is
little left for reordering to exploit, and at 64\textsuperscript{2} it \emph{adds} 21\%
DRAM.

\textbf{Tile membership.} Clustering the corpus so that articles sharing
rare descriptors fall in the same tile shrinks the union itself rather
than merely rescheduling it. This is the stronger form of the same idea,
and it made execution ordering a null - the clustered corpus already
supplies the locality the ordering was simulating. Ordering is retained
in this account as the experiment that identified the mechanism,
superseded by its own logic.

\textbf{Neuron-axis chunking.} A chunk is a contiguous slice of the
map's neurons. Splitting the neuron axis into C chunks means the kernel
sweeps a tile's feature columns against M/C of the M neurons at a time,
taking C passes to cover the whole map, so only one chunk's weights need
be resident at once. The co-resident working set is therefore
\texttt{union\ x\ (M/C)\ x\ 2} bytes - the tile's feature union, times
the neurons in one chunk, times two bytes per half-precision weight -
and is tunable against L2 at any map size. The registered prediction was
a residency threshold, a step in the hit rate as the working set crosses
the 72 MB cache. The measurement refutes the threshold and confirms the
mechanism: the hit rate climbed smoothly from 75.9\% to 97.9\% across C
from 1 to 16, with no step, so chunking buys partial residency and
gradual reuse. Occupancy was flat at 83\% throughout, which is the
diagnostic separating chunking from the tile sweep - two levers that
both look like "make the working set smaller" act on different
resources, one on cache and one on occupancy.

C topped out at 8 at 256\textsuperscript{2}. C = 16 continued to buy hit rate and cut DRAM
to 578 GB, and was slower: each chunk can only find the best match among
its own neurons, so the C partial results must be merged into one, and
that cost overtakes the memory saving.

\subsection{Vectorised loads, and a counter-intuitive
optimum}\label{vectorised-loads-and-a-counter-intuitive-optimum}

Vectorising the codebook loads as \texttt{\_\_half2} was the largest
single lever, 1.37-1.62x. It also moved the chunking optimum from 8 back
to 4, which \emph{raises} DRAM traffic and \emph{lowers} the hit rate.
\textbf{The optimum is not the configuration that maximises cache
statistics; it is the one that balances them against instruction cost.}
Any tuning account that selects on a cache metric rather than on time
would have taken the wrong branch here.

\subsection{Symmetrically tuning cuSPARSE and
SparseBin}\label{symmetrically-tuning-cusparse-and-sparsebin}

Tuning cuSPARSE began as fairness insurance and produced the programme's
most consequential single result.

Paper 1's cuSPARSE comparison spends three quarters of its time at 256\textsuperscript{2}
not in the sparse-dense product but in the argmin read-back that follows
it, streaming at 7.7\% of measured peak bandwidth. It assigns one thread
per article row, so warp-mates stride by M x 2 bytes and every warp
access spans 32 sectors: the same uncoalesced access pattern paper 1
identifies as the central error of node-major storage, committed inside
my own harness, on cuSPARSE's side of the comparison.

Rewritten as a warp-per-row reduction with vectorised loads it ran
\textbf{12.5x faster} at 256\textsuperscript{2}, 47.3 s to 3.8 s per epoch, taking DRAM
utilisation from 8\% to 96\%. At smaller maps the gains were 3.5x to
4.3x and the kernel reached 93-96\% of peak everywhere.

Two further cuSPARSE levers paid. Batch size is the structural analogue
of the tile: it sets the score block's footprint and therefore its cache
behaviour, and its optimum is map-size dependent in both directions,
with small maps wanting small batches and large maps the largest batch
that fits. Algorithm selection is the closest thing to a tile knob a
closed library offers, and \texttt{ALG3} proved a small-map win and a
large-map loss when swept alone, and a win at \textbf{every} map size
once crossed with batch size on the reordered corpus. That
non-additivity is why the programme switched to full factorials: a
one-factor-at-a-time sweep would have published a cuSPARSE optimum wrong
by 1.7x, in SparseBin's favour, and invisibly.

The conclusion this result selects was specified before it was run.
\textbf{The score block is affordable when it is read well}, not that
fusion was unnecessary. Against a coalesced read-back and a well-chosen
batch, fusion confers approximately no per-epoch advantage at 256\textsuperscript{2}. The
advantage of fusion is seen in the memory column of Table~\ref{tab:1}.

\section{Results}\label{results}

All timings were taken on one RTX 4090 against the frozen corpus.
Section 5.1 reports the comparison the parity rule governs; section 5.2
reports the two comparators it does not.

\subsection{Per epoch, both implementations
tuned}\label{per-epoch-both-implementations-tuned}

One epoch, seed 0, frozen corpus and split, RTX 4090. Median of three
replicates, spread $\le$1.5\% everywhere:

\begin{table}[htbp]
\centering
\small
\caption{Per-epoch time with both implementations tuned. One epoch, seed 0, frozen corpus and split, RTX 4090; median of three replicates.}
\label{tab:1}
\begin{tabularx}{\textwidth}{@{}llXlll@{}}
\toprule
edge & winner & configuration & s/epoch & peak memory & vs published \\
\midrule
32\textsuperscript{2} & cuSPARSE & warp argmax, ordered, ALG3, 64 MB batch & \textbf{0.24} & 2.5 GiB & 3.0$\times$ \\
64\textsuperscript{2} & SparseBin & vec2, tile 2 & \textbf{0.56} & 1.5 GiB & 6.4$\times$ \\
128\textsuperscript{2} & SparseBin & vec2, tile 2, clustered, C = 2 & \textbf{2.06} & 3.4 GiB & 10.1$\times$ \\
256\textsuperscript{2} & SparseBin & vec2, tile 4, clustered, C = 4 & \textbf{7.60} & 7.2 GiB & 5.6$\times$ \\
512\textsuperscript{2} & SparseBin & vec2, tile 4, clustered, C = 8 & \textbf{30.2} & 20.5 GiB & 6.2$\times$ \\
\bottomrule
\end{tabularx}
\end{table}

\subsection{The wider field: MedSOM and
somoclu}\label{the-wider-field-medsom-and-somoclu}

No new campaigns were run - every cell is the frozen record divided by
the winner table of Section~5.1.

\textbf{Uniform basis throughout: total wall clock $\div$ epochs, both
sides.}

\begin{table}[htbp]
\centering
\footnotesize
\caption{Margins over MedSOM and somoclu \cite{wittek2017}, published and after tuning. Uniform basis: total wall clock divided by epochs, both sides.}
\label{tab:2}
\begin{tabularx}{\textwidth}{@{}llrrrrr@{}}
\toprule
comparator & edge & \hd{comparator}{s/ep} & \hd{published}{s/ep} & \hd{published}{margin} & \hd{tuned}{s/ep} & \hd{tuned}{margin} \\
\midrule
MedSOM & 32\textsuperscript{2} & 37.63 & 2.393 & 15.7$\times$ & 0.31 & 121$\times$ \\
MedSOM & 64\textsuperscript{2} & 197.78 & 3.590 & 55.1$\times$ & 0.56 & 353$\times$ \\
MedSOM & \textbf{128\textsuperscript{2}} & 792.76 & 9.932 & \textbf{79.8$\times$} & 2.06 & \textbf{385$\times$} \\
MedSOM & 256\textsuperscript{2} & 3,209.88 & 42.25 & 76.0$\times$ & 7.60 & 422$\times$ \\
MedSOM & 512\textsuperscript{2} & OOM & - & - & 30.2 & - \\
somoclu & 32\textsuperscript{2} & 202.9 & 2.393 & 84.8$\times$ & 0.31 & 654$\times$ \\
somoclu & 64\textsuperscript{2} & 1,251.5 & 3.590 & 348.6$\times$ & 0.56 & 2,235$\times$ \\
somoclu & \textbf{128\textsuperscript{2}} & 6,168.5 & 9.932 & \textbf{621.1$\times$} & 2.06 & \textbf{2,994$\times$} \\
somoclu & 256\textsuperscript{2} & 27,685.0 & 42.25 & 655.3$\times$ & 7.60 & 3,643$\times$ \\
somoclu & 512\textsuperscript{2} & not run & - & - & 30.2 & - \\
\bottomrule
\end{tabularx}
\end{table}

\textbf{Matched work} is the first paper's Section~5.5 definition: a fixed 20
epochs over the full corpus (26,912,934 rows per epoch), the same 90/10
split, the same map, total wall $\div$ epochs. The tuned denominators are
one-epoch runs, three replicates, spread $\le$1.5\%, with KL instrumentation
left in - so they are conservative.

\textbf{128\textsuperscript{2} is quoted throughout the first paper, and it is not the
best edge.} At 256\textsuperscript{2} the margins are larger, 422$\times$ and 3,643$\times$; quoting
128\textsuperscript{2} keeps continuity with the first paper's published figures rather
than selecting the maximum.

\textbf{Neither MedSOM nor somoclu was retuned.} The parity rule was
scoped to the two sparse implementations that compete on the same
design, cuSPARSE and SparseBin; MedSOM and somoclu ran at their
published or default configurations, so these two margins rest on weaker
ground than those in Table~\ref{tab:1}.

\begin{figure}[htbp]
\centering
\includegraphics[width=0.86\textwidth]{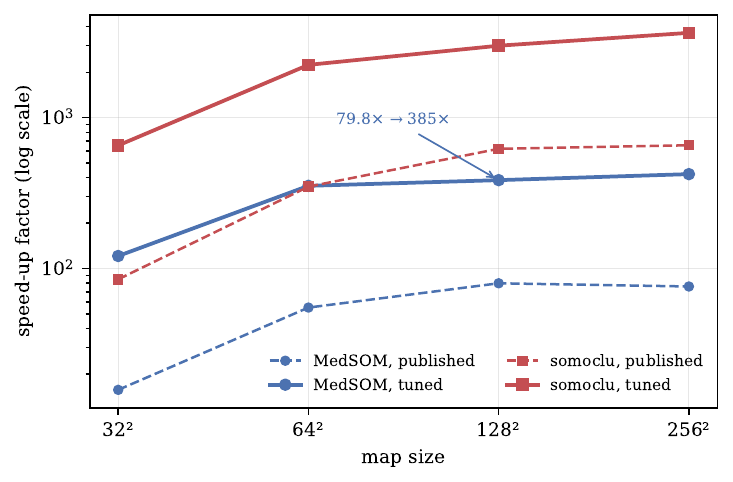}
\caption{Speed-up over MedSOM and somoclu, as published and after tuning, by map size (log scale). The MedSOM pair is a GPU-to-GPU comparison; the somoclu pair is against a multicore-CPU library. Neither comparator was retuned.}
\label{fig:margins}
\end{figure}

\subsubsection{A basis inconsistency in paper 1, which this paper does
not
propagate}\label{a-basis-inconsistency-in-paper-1-which-this-paper-does-not-propagate}

Paper 1's abstract and Section~5.5 state \ensuremath{\sim}82$\times$ against MedSOM and
621$\times$ against somoclu in the same sentence. Those two figures do not
share a denominator. The 621$\times$ divides by the total-wall per-epoch figure
of 9.932 s; the 82$\times$ divides by a training-loop figure of
\ensuremath{\sim}9.67 s that excludes initialisation and evaluation
amortisation. On the uniform basis used in the table above, the MedSOM
margin is \textbf{79.8$\times$}, a discrepancy of under 3\%. The title rounds
that to 80$\times$.

This paper prints the recomputable figure. Where paper 1's number is
referred to, it is attributed - "82$\times$ as published" - rather than
reproduced as though it were derived here. A paper arguing that
comparisons should be checkable cannot carry a headline figure that does
not recompute from its own artefact.

\textbf{This is also a correction item for paper 1's journal version.}
It changes no conclusion; it is a denominator that should be stated once
and used twice.

\section{The performance ceiling}\label{the-performance-ceiling}

\begin{table}[htbp]
\centering
\small
\caption{Limiter progression across the programme. The shipped kernel is the first configuration to press a hardware roof rather than sit beneath all of them.}
\label{tab:3}
\begin{tabularx}{\textwidth}{@{}llXl@{}}
\toprule
& first addendum (tile 16) & scalar optimum & shipped \\
\midrule
registers/thread & 121 & 46 & 46 \\
achieved occupancy & 33\% & 83\% & 66\% \\
DRAM & 45\% of peak & 15\% & 41\% \\
L2 & 0.7 TB/s (14\%) & 2.5 TB/s (50\%) & \textbf{3.85 TB/s (77\%)} \\
warp-issue & 30\% & 61\% & 64\% \\
busiest pipe & ALU 21\% & LSU 61\% & LSU 49\% \\
limiter & latency, under every roof & issue/LSU, under every roof & \textbf{at the L2 roof} \\
\bottomrule
\end{tabularx}
\end{table}

The shipped kernel was the first configuration in the sequence to press
a hardware bandwidth ceiling rather than sit beneath all of them, and
the same signature appeared at 512\textsuperscript{2}. That bounds any future lever at
roughly \textbf{1.3$\times$} - measured as 1.29$\times$ whole-epoch at 512\textsuperscript{2}, 1.19$\times$ at
32\textsuperscript{2}.

\begin{figure}[htbp]
\centering
\includegraphics[width=0.86\textwidth]{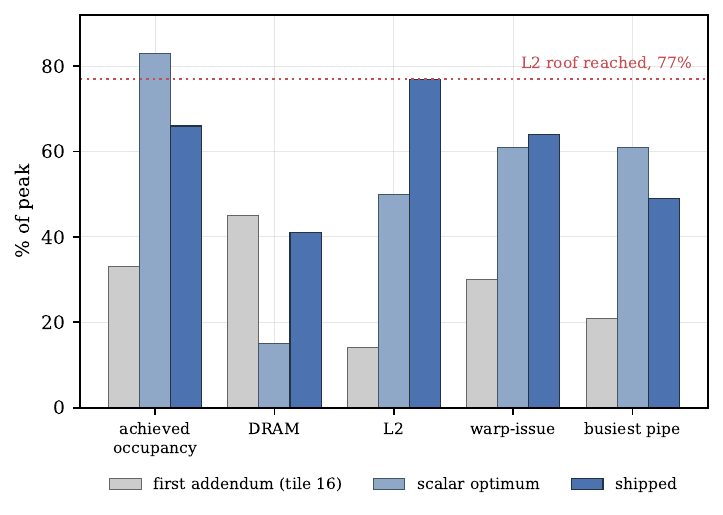}
\caption{Limiter progression across the programme. The shipped kernel is the first configuration to press a hardware ceiling rather than sit beneath all of them.}
\label{fig:limiters}
\end{figure}

\subsection{Composition of the epoch
wall-time}\label{composition-of-the-epoch-wall-time}

The search was 98.7-99.5\% of the epoch wall-time when the programme
began. Making it roughly six times faster raises the question of whether
the rest of the epoch has become material. It has not, and the direction
is the opposite of what one would guess. One instrumented epoch at the
shipped configuration, free-running clocks, kernel-time sums; the device
total matched the trainer's own wall to within 0.02 s at every map size,
so host-side work during the epoch is nil.

\begin{table}[htbp]
\centering
\footnotesize
\caption{Composition of one instrumented epoch at the shipped configuration, in seconds.}
\label{tab:4}
\begin{tabularx}{\textwidth}{@{}lrrrrrrrr@{}}
\toprule
edge & wall & \hd{fused}{BMU} & \hd{stopping}{metric} & blur & scatter & \hd{write-}{back} & norms & \hd{BMU}{share} \\
\midrule
32\textsuperscript{2} & 0.32 & 0.236 & 0.033 & 0.001 & 0.033 & 0.001 & 0.001 & 73.7\% \\
64\textsuperscript{2} & 0.56 & 0.470 & 0.043 & 0.003 & 0.034 & 0.003 & 0.002 & 83.9\% \\
128\textsuperscript{2} & 2.05 & 1.817 & 0.122 & 0.035 & 0.037 & 0.016 & 0.002 & 88.6\% \\
256\textsuperscript{2} & 7.56 & 6.915 & 0.358 & 0.145 & 0.035 & 0.064 & 0.014 & 91.5\% \\
512\textsuperscript{2} & 29.98 & 28.175 & 0.820 & 0.592 & 0.035 & 0.260 & 0.065 & 94.0\% \\
\bottomrule
\end{tabularx}
\end{table}

Under the quantisation-error stop the stopping-metric column disappeared
and nothing else moved beyond noise, taking the search's share to 83.3 /
90.4 / 94.2 / 96.0 / 96.5\%.

\textbf{The update's two components scale as the design predicts, and in
opposite directions.} The scatter into accumulators is O(N.nnz) and
independent of map size: 0.033 to 0.037 s at every rung, across a
256-fold range of neuron counts. The blur is the O(M) term and grows
accordingly, 0.001 to 0.592 s. Their sum grows far more slowly than the
search does, so \textbf{the search's share rises with map size.} The
single-number epoch was least defensible at the \emph{smallest} maps,
where non-search work is 26\% of the epoch, and most defensible at the
largest, where it is 3.5\%.

\textbf{The whole-epoch ceiling follows.} With the search at 96.5\% of
the epoch at 512\textsuperscript{2} and the L2 roof bounding search work at about 1.3x,
the whole-epoch bound is roughly \textbf{1.29x}. At 32\textsuperscript{2}, where
non-search work is a quarter of the epoch, it is about \textbf{1.19x}.
The update phase costs about one point of the kernel ceiling at large
maps and fifteen at small ones, which is the honest bound on any further
work of the kind reported here.

\begin{figure}[htbp]
\centering
\includegraphics[width=0.86\textwidth]{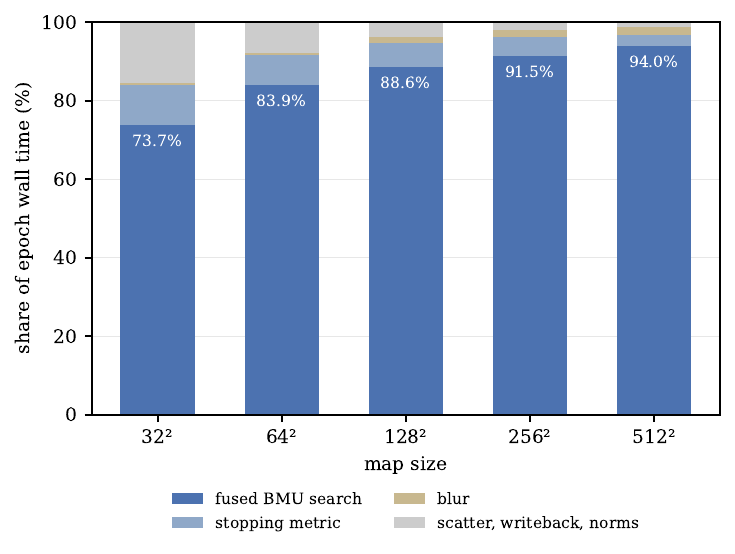}
\caption{Composition of one instrumented epoch at the shipped configuration. The search's share rises with map size, so the single-number epoch is least defensible at the smallest maps and most defensible at the largest.}
\label{fig:epoch}
\end{figure}

\subsection{The unpulled levers}\label{the-unpulled-levers}

A number of levers were considered but not pulled.

\begin{table}[htbp]
\centering
\small
\caption{The unpulled levers and why each was left.}
\label{tab:5}
\begin{tabularx}{\textwidth}{@{}L{0.75}L{0.55}L{1.70}@{}}
\toprule
lever & status & why \\
\midrule
\texttt{bmu1-only} & scoped out, twice & compute/latency lever, reduces no L2 bytes $\rightarrow$ capped $\lesssim$1.3$\times$ by construction; and second-best feeds topographic error \cite{kaski1996}, which both stopping rules gate on at convergence \\
shared-memory staging, \texttt{cp.async} & scoped out & compute/latency levers; neither reduces bytes crossing L2 \\
nnz binning & scoped out & redistributes instruction work, not L2 traffic \\
\textbf{L1 carveout} & \textbf{measured null} & the one lever whose mechanism aims at the binding resource - so it was measured rather than argued away \\
\bottomrule
\end{tabularx}
\end{table}

Each streaming multiprocessor has a fixed pool of on-chip memory that is
divided between the L1 cache and the shared memory a kernel can address,
and the split can be requested per kernel. Since a larger L1 share
absorbs read requests before they reach L2, this is the one untried
lever whose mechanism acts on the binding resource itself. The split was
swept from the driver's own choice to both extremes, one epoch per cell,
with quantisation error checked identical throughout:

\begin{table}[htbp]
\centering
\small
\caption{L1 / shared-memory split sweep on the shipped kernel, seconds per epoch.}
\label{tab:6}
\begin{tabular}{@{}llllll@{}}
\toprule
edge & driver default & max L1 & 25\% shared & 50\% shared & max shared \\
\midrule
128\textsuperscript{2} & \textbf{2.07 s} & 2.12 & 2.24 & 2.24 & 2.24 \\
256\textsuperscript{2} & \textbf{7.55 s} & 8.33 & 7.66 & 7.71 & 7.72 \\
512\textsuperscript{2} & \textbf{30.11 s} & 32.20 & 30.33 & 30.44 & 30.47 \\
\bottomrule
\end{tabular}
\end{table}

The driver's own choice was fastest at every map size. Forcing maximum
L1, the setting the mechanism argued \emph{for}, was the \textbf{worst}
of the five, by 2.4\%, 10.3\% and 6.9\%. Requesting any fixed split
removes the driver's freedom to vary it, and the kernel uses so little
shared memory that the default already favoured L1. The lever is a null,
and the fact that it fails on the wrong side of the prediction
strengthens the case for leaving it alone.

\section{Predictions and nulls}\label{predictions-and-nulls}

\subsection{Registered predictions}\label{registered-predictions}

Seven predictions bearing on the results reported here were registered
before the measurement that tested them.

\begin{table}[htbp]
\centering
\footnotesize
\caption{Predictions registered before the measurements that tested them.}
\label{tab:7}
\begin{tabularx}{\textwidth}{@{}lL{1.05}L{0.95}@{}}
\toprule
\# & Registered & Outcome \\
\midrule
1 & Ordering reduces L2 traffic \textgreater=20\% at tile 2, 128\textsuperscript{2} & \textbf{Miss} on L2 bytes (-0.8\%); hit on DRAM (-40\%) under the metric reframed before measurement \\
2 & At 128\textsuperscript{2}, ordering makes the working set resident: hit \textgreater95\%, DRAM down \textgreater2x & \textbf{Miss} - 82.4\%, 1.66x \\
3 & At 256\textsuperscript{2}, ordering moves the tile optimum down and DRAM below saturation & \textbf{Hit} \\
4 & Chunking tunes residency, with a threshold as the working set crosses L2 & \textbf{Split} - mechanism confirmed, threshold refuted; smooth partial residency \\
5 & The 256\textsuperscript{2} dead-unit anomaly does not reproduce at the new optimum & \textbf{Hit} \\
6 & Hot-column persistence helps cuSPARSE materially more & \textbf{Refuted} - helped neither \\
7 & Ordering causes cuSPARSE's epoch inflation & \textbf{Refuted} - one epoch; the cause was a protocol error in that campaign \\
\bottomrule
\end{tabularx}
\end{table}

Predictions 2 and 6 were the informative failures. Prediction 2
conflated a resident \emph{codebook} with a resident \emph{working set}:
one stays put, the other turns over, and cross-window eviction eats the
margin. Prediction 6 could have been reasoned out in advance -
cuSPARSE's argmin reads the score block, not the codebook, so pinning
hot codebook columns was never going to reach its bottleneck.

\subsection{Nulls and refutations}\label{nulls-and-refutations}

\begin{table}[htbp]
\centering
\footnotesize
\caption{Nulls and refutations.}
\label{tab:8}
\begin{tabularx}{\textwidth}{@{}L{0.85}L{1.15}@{}}
\toprule
Lever & Outcome \\
\midrule
\texttt{cusparseSpMM\_preprocess} and persistent descriptors & Null at every map size and algorithm. The defect was real; its cost was invisible at one-epoch granularity \\
Hot-column L2 persistence, both implementations & Null on both sides, under 1\% \\
Frequency-ordered feature relabel & Timing null, \textless=1.5\%; retained as substrate \\
Reduced-precision score block & Null by construction - already half precision in every published comparison \\
CSC tile transpose & Retired by measurement: its target fell from 19\% to 10\% of the instruction stream when the tile narrowed. Ceiling \ensuremath{\sim}1.1x \\
\texttt{\_\_launch\_bounds\_\_} register caps & No win; spills convert register pressure into cache and memory traffic \\
Second-best-unit tracking removed & Correct lever, wrong bottleneck at the published tile; superseded \\
Chunking at C = 16 & Fits with a two-buffer merge and buys nothing \\
Armed Kaski-Lagus evaluation & Degenerate. The trigger fired at epoch 3 in every run and never disarmed, because the improvement series is non-monotone; no choice of threshold fixes it \\
Execution ordering & Superseded by clustered membership, which supplies the same locality \\
\texttt{CUSPARSE\_SPMM\_CSR\_ALG1} & 4-6x slower than the default everywhere. A hazard, recorded as a warning \\
\bottomrule
\end{tabularx}
\end{table}

\section{The cost of symmetry}\label{the-cost-of-symmetry}

The cost of symmetrically tuning both algorithms was roughly 11.5
GPU-hours and 16 hours of implementation on this side; 4.6 and 9 on
cuSPARSE, for the tuning programme alone. The validation,
protocol-correction and packaging campaigns that followed added about
another 18 GPU-hours, split roughly 55:45 across the two sides.

\textbf{The gap is real and is not a fairness failure.} cuSPARSE's core
is a closed library, so its one tunable kernel received its retune and
had no vectorisation left to receive, while SparseBin has more of its
own code to tune \emph{because it is its own code}. That a bespoke
kernel remains tunable where a library call does not is part of the
finding rather than an apology for it. Every lever with an analogue on
both sides was run on both.

The ratio matters more than the total: the parity work was about a third
of the effort, and most of that third was spent once, on understanding
cuSPARSE well enough to know what to offer it.

\section{Availability of code and
data}\label{availability-of-code-and-data}

Everything needed to repeat the measurements in this paper is public.
The repository \texttt{sparsesom-tuning} holds the tuning and validation
campaign: the scripts exactly as they were run, the per-phase timings
behind every table above, the profiler captures \cite{nsight} they were
derived from, and a record of which binary and source revision produced
each campaign. The corpus itself is deposited separately and openly, so
a reader can start from the same 29.9 million documents rather than
approximating them.

The repository is \texttt{sparsesom-tuning} at release \texttt{v2.0},
archived at Zenodo concept DOI \texttt{10.5281/zenodo.22245712}. A
SHA-256 manifest covers every file, and the record marks which campaigns
are the corrected runs. One campaign was withdrawn during the programme
after a protocol error was found in it; it is retained in the repository
and labelled as withdrawn rather than deleted.

The repository was tested as a stranger would use it before release:
cloned fresh into a clean directory, built from the committed scripts
alone, and used to re-derive the headline table from the corpus at its
public DOI. Seven of ten per-epoch cells reproduced exactly, and three
carry each binary's first-launch just-in-time compilation, documented
with warm-up guidance.

That exercise earned its place twice. The first attempt failed on a
virgin configure, where the default host compiler's standard-library
types break the CUDA front end - invisible in a campaign whose build
trees were configured months earlier. Both defects it found were
packaging defects and no measurement moved, which is the outcome one
hopes for and cannot assume.

\section{Related work}\label{related-work}

\subsection{Symmetrical tuning}\label{symmetrical-tuning}

\textbf{Automated tuning.} Search-based autotuners - ATLAS in the
dense-linear-algebra tradition, and general frameworks such as
OpenTuner, CLTune and Kernel Tuner - exist precisely to relieve an
author of hand-tuning, and an obvious question is why this programme was
run by hand. Two reasons, both visible in the results. First, the levers
interact non-monotonically: vectorisation moved the chunking optimum
from 8 back to 4, and moved it in the direction that \emph{raises} DRAM
traffic and \emph{lowers} hit rate (Section~4). A search that optimises one
parameter at a time converges to a local optimum and reports it as an
answer, and the parameter that has to move to escape it is not the one
being searched. Second, and more to the point of this paper, an
autotuner tunes the parameters you expose to it. On cuSPARSE's side
those are the library's algorithm selector, its batch size, and the
precision of its score block - which is a real search space, and was
searched, but it is not the same act as asking what cuSPARSE's
bottleneck actually is and finding that its argmin read-back was
uncoalesced.

\textbf{Roofline and limiter analysis.} The ceiling argument in Section~6 rests
on the roofline model \cite{williams2009}, and specifically on its
instruction-roofline formulation for GPUs \cite{ding2019}, which the companion
profiling addendum uses directly. The contribution here is not
methodological: it is that a tuning programme was run until the kernel
reached a roof, and that the roof is then used to bound what remains
rather than to characterise what was achieved.

\textbf{Self-organizing maps at scale.} Kohonen's formulation \cite{kohonen2013},
and the parallel implementations that followed it - somoclu as the
multicore-CPU reference \cite{wittek2017}, and MedSOM as the CUDA implementation
behind our earlier atlases \cite{amos2024a,amos2024b}. This is context rather than
contest; Section~5.3 reports the margins and states plainly which of these the
parity rule was applied to.

\section{Conclusion}\label{conclusion}

Tuning the best-matching-unit search through four standard levers made
it 5.6-10.1$\times$ faster per epoch than the configuration previously
published. The comparison it is set against is not the one the earlier
paper used: cuSPARSE received every applicable lever from the same
programme and became 2-3$\times$ faster in the process. The margin survives
that improvement, which is the only reason it is worth reporting.

The search stopped at a ceiling rather than at exhaustion. The tuned
kernel pressed the L2 bandwidth roof at 77\% of peak while every other
unit sat at 40-65\%, which bounds any further lever on this device at
roughly 1.3$\times$ - measured as 1.29$\times$ whole-epoch at 512\textsuperscript{2} and 1.19$\times$ at 32\textsuperscript{2}.
Every lever left untried is accounted for against that bound, and the
one whose mechanism aimed at the roof itself was measured rather than
argued away.

This is a single-device study. The values here are specific to one
consumer card, and I expect the balance to move on a part with different
cache and bandwidth proportions. What transfers is not the tile size or
the chunk count. It is the observation that a comparison's denominator
is a variable, that it is worth two to three times in this case, and
that reporting it costs about a third of the effort already being spent.


\end{document}